\documentclass[letterpaper, 10 pt, conference]{ieeeconf}  

\IEEEoverridecommandlockouts                              

\usepackage{graphicx}

\usepackage{amsthm}
\usepackage{amsmath} 
\usepackage{amssymb}  
\usepackage{amsthm}
\usepackage{bm}
\usepackage{xcolor}
\usepackage{float}
\usepackage{algorithm}
\usepackage{algorithmic}
\usepackage[colorinlistoftodos]{todonotes}
\usetikzlibrary{calc,decorations.pathreplacing}
\usepackage{optidef}
\usepackage{subcaption}
\usepackage{cite}
\usepackage[colorlinks=true, linkcolor=black, citecolor=black, urlcolor=black]{hyperref}
\usepackage[normalem]{ulem}

\usepackage{siunitx}
\usepackage{layouts}
\usepackage{orcidlink}
\usepackage{eso-pic} 
\usepackage{fancyhdr}
\fancypagestyle{firstpage}{
  \fancyhf{}
  \fancyfoot[C]{\scriptsize \copyright~2026 IEEE. Personal use of this material is permitted. Permission from IEEE must be obtained for all other uses, in any current or future media, including reprinting/republishing this material for advertising or promotional purposes, creating new collective works, for resale or redistribution to servers or lists, or reuse of any copyrighted component of this work in other works.}

}

\usepackage[normalem]{ulem}

\newcommand{\nvec}{\bm} 
\newcommand{\nmat}{\bm} 
\newcommand{\transposed}{^\top}
\newcommand{\transform}[3]{#1_{#2,#3}} 
\newcommand{\nset}{\mathcal}

\definecolor{tristancolor}{rgb}{0.5,0.5,0.75}

\title{\LARGE \bf
Real-Time and Accurate Collision-Free Teleoperation via Differentiable Constraint-Based Trajectory Planning
}

\author{Max Grobbel\textsuperscript{*}$^{1\orcidlink{0009-0009-7110-9998}}$, Tristan Schneider\textsuperscript{*}$^{1\orcidlink{0000-0002-4804-3902}}$, Daniel Flögel$^{1}$, Sören Hohmann$^{2}$
\thanks{$^{1}$Max Grobbel, Tristan Schneider and Daniel Flögel are with FZI - Forschungszentrum Informatik,  Karlsruhe, Germany
        {\tt\small grobbel@fzi.de}}%
\thanks{$^{2}$Sören Hohmann is with the Department of Electrical Engineering, Karlsruhe Institute of Technology, Karlsruhe, Germany}
\thanks{* equal contribution}
}

\begin{document}

\maketitle
\thispagestyle{firstpage}
\pagestyle{empty}

\begin{abstract}

In teleoperation, the human operator typically controls only the end-effector pose, which often leads to self-collisions of the manipulator and collisions with environmental obstacles, since joints and links are not controlled individually. A common strategy to mitigate this issue is to enhance the operator’s input using optimal-control-based trajectory planning. As derivative-based solvers require differentiable constraints, existing approaches either approximate robots and obstacles with spheres, reducing geometric accuracy, or approximate derivatives, degrading convergence and increasing computation times. We address these limitations by adapting a recent formulation of differentiable collision-avoidance constraints, based on duality in convex optimization, to the teleoperation setting. The robot is approximated with capsules and the environment with polytopes. We compare the resulting trajectory planning method against state-of-the-art techniques in simulation with varying numbers of obstacles and evaluate it on a UR5e manipulator in a real-world teleoperation test. Results show that our approach achieves lower computation times while enabling more accurate obstacle modeling, leading to smoother and collision-free end-effector teleoperation.

\end{abstract}

\section{Introduction}
Teleoperation of robotic manipulators is an upcoming field of research with classical applications in hazardous areas like nuclear power plant dismantling or in hard-to-reach places such as space applications \cite{hokayem2006a}. Bringing human experts, e.g., surgeons, into remote places or data collection for machine learning applications are more modern applications of teleoperation \cite{siciliano2016a, lichiardopol2007a, wu2024a}.

A central component of teleoperation systems is the input device, which can be broadly classified as haptic or non-haptic. Haptic devices provide rich feedback and often replicate the kinematic chain of the controlled robot, enabling the operator to manipulate not only the end-effector but also individual robot links. However, haptic input devices are expensive, preventing widespread applications. In contrast, non-haptic input devices, e.g., camera tracking or motion tracking systems, are available for lower prices, but provide less feedback to the user, often leading to lower performance. Also, and more importantly, non-haptic input devices only allow the control of the end-effector pose. As indicated in \cite{wu2024a}, this limitation leads to a larger number of collisions of the manipulator with itself and the environment, since the operator must implicitly account for the robot’s full inverse kinematics.

\begin{figure}[ht!]
    \vspace{5pt}
    \centering
    \includegraphics[width=1\linewidth]{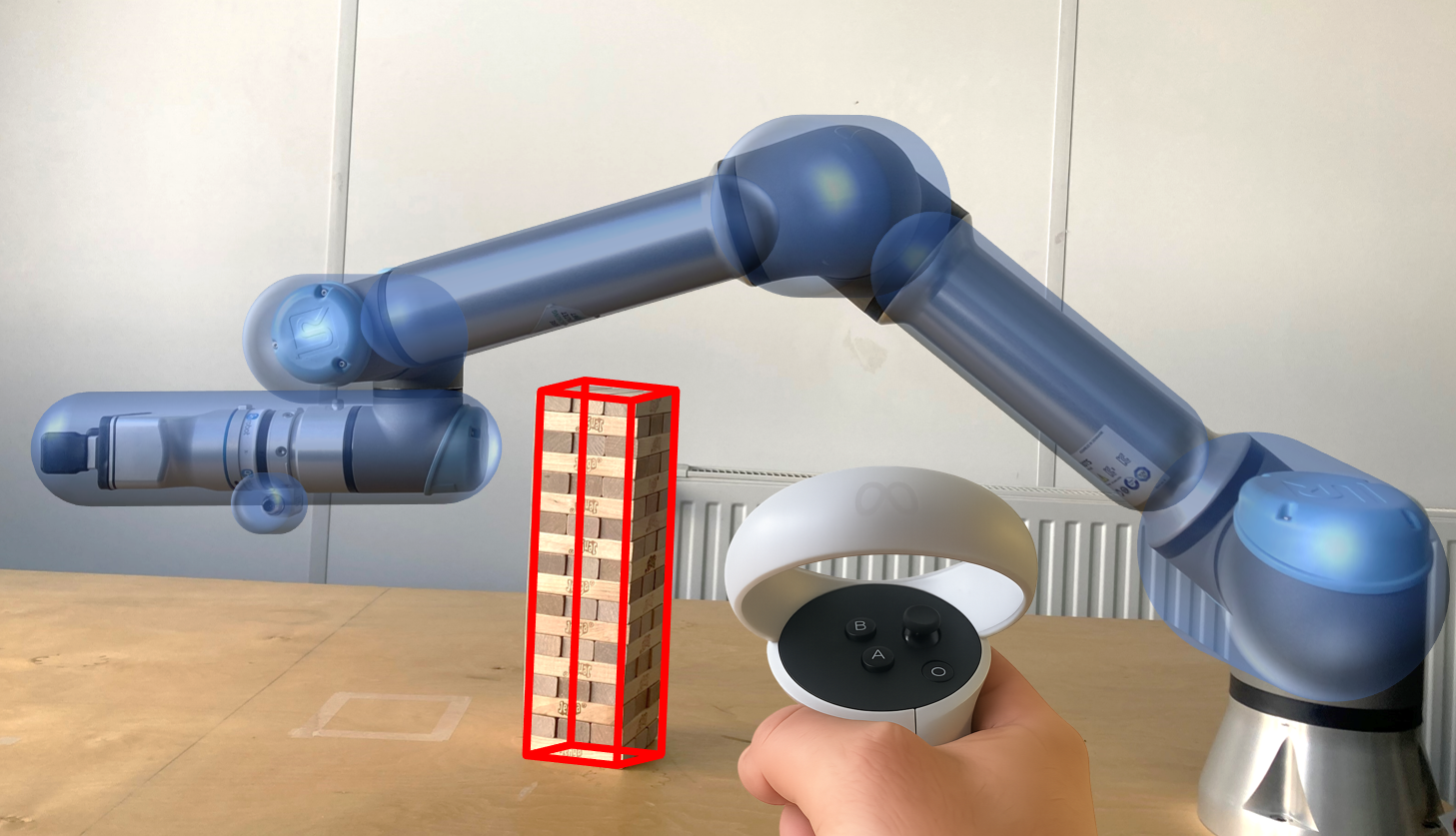}
    \caption{Teleoperated robotic manipulator. For collision avoidance, the manipulator is approximated by capsules, while environmental obstacles are modeled as polytopes.}
    \label{fig:UR_capsules}
    \vspace{-5pt}
\end{figure}

This shows the necessity to support the human operator in teleoperation scenarios, especially with collision avoidance. A commonly used strategy to assist the human operator in teleoperation setups is the application of shared control algorithms \cite{li2023b}. Those can be categorized into algorithms that generate a single set point for low-level controllers (e.g. \cite{rakita2017a}) and approaches, which use the human input as a tracking reference and plan slightly adjusted trajectories, either in task space or in joint space.

Three widely adopted approaches to trajectory planning in robotics are based on sampling methods \cite{elbanhawi2014a}, optimal control formulations \cite{zhao2025a}, and learning-based algorithms \cite{wang2021c}. While sampling- and learning-based methods generally lack inherent mechanisms for enforcing constraints, optimal control approaches explicitly exploit system dynamics, naturally accommodate state and environmental constraints, and allow for the formulation of multiple objectives, provided these objectives are expressed in a differentiable form—a requirement stemming from their reliance on derivative-based optimization solvers. Moreover, using forward kinematics within the optimal control formulation implicitly resolves the inverse kinematics and generates task-space motions \cite{faulwassertimm2017a}.

Trajectory planning in teleoperation can be regarded as a specialized instance of robotic trajectory planning. The presence of a human operator in the loop introduces additional requirements, most notably real-time capabilities to ensure rapid responsiveness and the ability to adapt to time-varying reference trajectories arising from stochastic human input. Optimal control–based approaches are well suited to these demands, as they combine dynamic models with constraint handling and real-time optimization. Consequently, such methods have been widely applied in teleoperation contexts \cite{lima2024a, muchacho2023a, rubagotti2019a, hu2021a}.

Collision between the robot and its environment, including self-collisions, is avoided by enforcing a minimum distance between bodies. Optimization-based methods require differentiable constraint functions, which poses a challenge for collision avoidance: the minimum-distance computation between two bodies is itself a minimization problem and in general non-differentiable. Existing teleoperation approaches address this issue either by approximating robots and obstacles with numerous spheres \cite{sundaralingam2023a}, which leads to a trade-off between approximation accuracy and computational cost, or by employing non-differentiable distance computations with subsequently approximated gradients, as in \cite{hu2021a}.

Recent research has introduced an algorithm for differentiable collision constraints for convex bodies \cite{zhang2021a}. This approach was refined by \cite{dietz2023a}, who reduced the computational demand and applied it to a path planning problem for mobile platforms, though the computation time remained above one second, limiting real-time use. The underlying idea is to reformulate the minimum-distance computation between two bodies—originally a nested, non-differentiable optimization—into a dual maximization problem, yielding a set of differentiable inequality constraints that enforce minimum separation.

The main contribution of this work is an optimal control formulation for teleoperation that contains collision-avoidance constraints for accurate geometric robot and environment models. Specifically, we extend the duality-based approach of \cite{dietz2023a} by introducing capsules as additional geometric primitives and approximating the geometry of an industrial robot manipulator with capsules (\autoref{fig:UR_capsules}). The resulting constraints are directly embedded into a model predictive trajectory planner, yielding a real-time capable method that ensures both self-collision avoidance and safe interaction with polytope-shaped environmental obstacles.

We evaluate this approach in terms of computational efficiency and tracking accuracy, comparing it against state-of-the-art teleoperation methods in both simulation benchmarks and physical robot experiments.

\section{Related Work}
Collision avoidance with the environment and with the robot itself is essential for teleoperation tasks with motion tracking devices, since the operator only controls the end-effector pose. 
The real-time requirements and the time-dependent tracking objective make teleoperation a distinct subproblem of trajectory planning for robotic manipulators. In the following, we first provide an overview of collision-avoidance formulations within optimal control for robotic manipulators in general, before reviewing approaches tailored to the specific requirements of teleoperation.

\subsection{Collision avoidance for robotic manipulators}
Collision-avoidance for manipulators in optimal control can be divided into five categories: joint-space formulations, and task-space formulations based on spheres, witness points, dual collision-multipliers, or other approximation-based methods such as finite-difference or scaling techniques.

The first group formulates collision constraints in joint space \cite{muchacho2023a}, where each individual joint angle is restricted to remain within a predefined interval. While such constraints are straightforward to incorporate into optimization problems, the decoupled consideration of joint angles leads to a conservative approximation of the free space.

A common approach for task-space formulation is based on the approximation of all relevant bodies through spheres. Depending on the geometry and the desired accuracy, the number of spheres can become large. The distance between two spheres is computed by the Euclidean distance between the two center points and the subtraction of both radii. The spheres of all possible collision pairs need to be checked for collision. Those distance functions are differentiable with respect to the robot configuration and can therefore be directly integrated into an optimization problem \cite{rubagotti2019a, lima2024a, stoyanov2018a}.

Another approach is based on the widely used trajectory optimization algorithm of Schulman \cite{schulman2013a}. The collision avoidance relies on fast distance-computation techniques, like the Gilbert-Johnson-Keerthi (GJK) algorithm, originating in the domain of video games. These algorithms return both the minimum distance between two bodies and the corresponding closest points---also referred to as witness points \cite{haffemayer2024a}. Schulman's method approximates the derivative of the distance with respect to the robot configuration (i.e., joint angles of a manipulator), resulting in a linear, differentiable approximation of the distance. The witness points are assumed to remain fixed in the local position on the bodies. The optimization problem is solved with a sequential-quadratic-programming (SQP) approach, giving the opportunity to linearize around a new solution during the SQP loop. However, the assumption of fixed witness points can lead to inaccuracies for larger configuration changes, limiting the method’s robustness in complex teleoperation scenarios.

Finite differences provide another option for estimating the derivative of non-differentiable algorithms \cite{kramer2020a} with the downsides of numerical differentiation. An alternative strategy avoids explicit distance computation by expanding all relevant objects with a scaling factor \cite{tracy2023a}. With a scaling factor larger than one, the objects are not in collision. The resulting equations are differentiable with respect to the robot configurations. As a downside, this method only enforces collision avoidance without respecting a prescribed minimum distance. 

These limitations motivate more principled formulations, such as dual collision-multiplier approaches, which provide differentiable constraints while explicitly enforcing minimum distance requirements \cite{zhang2021a}. In this method, the distance computation, which is a minimization problem by definition, is transformed into a system of inequalities with the exploitation of the dual problem. While the initial formulation was validated for mobile platforms, subsequent work has extended the approach to robotic manipulators \cite{he2025, lin2025}, albeit without achieving real-time performance. A recent adaptation \cite{dietz2023a} demonstrated promising improvements regarding the computation time for path planning of mobile robots, supporting ellipsoids and polytopes as geometry approximations for the relevant obstacles.

\subsection{Collision avoidance in teleoperation for robotic manipulators}
 Although some methods for collision avoidance in teleoperation without optimal control exist (e.g. \cite{wang2015a}), most approaches rely on optimal control, typically using geometric modeling through spheres or witness points.

For example, \cite{rubagotti2019a} approximate both the robot and environmental obstacles with spheres. They approximate every robot link with three spheres, which leads to a very conservative behavior. The optimization problem is solved using an SQP scheme and contains constraints to prevent collision with the environment, whereas self-collision is not addressed. 

In contrast, \cite{hu2021a} integrate both environmental and self-collision constraints into the optimization problem using the witness-point approach. This method can handle arbitrary convex geometries and leverages the Flexible Collision Library (FCL) to compute distances and witness points. The resulting optimization problem is solved with an SQP scheme. A key limitation, however, is the linear approximation of the distance function and its gradient, possibly leading to convergence issues and thus requiring a larger number of SQP iterations, leading to long computation times.

To date, no teleoperation approach has provided differentiable collision-avoidance constraints together with accurate geometric approximations that can be integrated into the optimization problem, while simultaneously ensuring real-time feasibility and handling both self-collision and environmental collision.

\section{Duality Based Collision Constraints}
To derive differentiable collision-avoidance constraints, we first introduce geometric approximations of the manipulator and the surrounding obstacles. The distance computation between two bodies---formulated as a minimization problem by definition---is then reformulated as a system of inequalities \cite{dietz2023a}. These can subsequently be integrated into the optimal control problem (OCP) as collision avoidance constraints.

\subsection{Geometric Approximation}
All bodies are mathematically formulated as subsets $\nset B \subset \mathbb R^3$ of the three-dimensional work space. A convex polytope $\nset P$ can be represented as the convex hull of its vertices, which is given by
\begin{equation}
    \nset P = \left\{\nmat V \nvec\varphi\ \vert\ \nvec\varphi\in\mathbb R^{N_\text{v}}, \nvec 1\transposed\nvec\varphi = 1, \nvec\varphi \geq \nvec 0\right\}, 
    \label{eq:polyeder_menge}
\end{equation}
with $\nmat V \in \mathbb R^{3\times N_\mathrm{v}}$ being the matrix that contains the $N_\mathrm{v}$ vertices as its columns~\cite{dietz2023a}. A capsule $\nset C$ is defined as the Minkowski sum
\begin{equation}
    \nset C = \left\{\nvec l + \nvec\delta\ \vert\ \nvec l \in \nset L, \nvec\delta \in \mathbb R^3, \left\lVert \nvec\delta \right\rVert \leq r \right\}
\end{equation}
of a line segment $\nset L \subset \mathbb R^3$ and a closed ball. This results in a set of all points that have an Euclidean distance to the line segment that does not exceed a certain radius $r\in \mathbb R$. The line segment is formulated as a convex polytope with two vertices and the vertex matrix $\nmat V_{\nset L} = \begin{bmatrix}\nvec v^{(1)} & \nvec v^{(2)}\end{bmatrix}$.

The robotic manipulator is geometrically approximated using multiple capsules (\autoref{fig:UR_capsules}). To ensure that all potential collisions are accounted for, the real geometry of the robot links needs to be contained completely in the capsule-based collision model. Each capsule is associated with one link of the kinematic chain. The configuration-dependent locations of the center line end points are given by
\begin{equation}
    \begin{bmatrix}
        \nvec v^{(1)} & \nvec v^{(2)}
    \end{bmatrix}
    = \transform{\nmat R}{0}{i}
    \begin{bmatrix}
        \tilde{\nvec v}^{(1)} & \tilde{\nvec v}^{(2)}
    \end{bmatrix}
    + \transform{\nvec p}{0}{i} \nvec 1\transposed,
    \label{eq:vertices_transformation}
\end{equation}
where $\transform{\nmat R}{0}{i}$ and $\transform{\nvec p}{0}{i}$ are the rotation matrix and translation vector describing the transformation of the corresponding $i$-th link with respect to a fixed global coordinate system. The vectors $\tilde{\nvec v}^{(1)}$ and $\tilde{\nvec v}^{(2)}$ contain the constant coordinates of the end points in the local coordinate system of the link and $\nvec 1 = \begin{bmatrix}
    1 & \dots & 1
\end{bmatrix}\transposed$ is a vector of ones.

Obstacles present in the environment are modeled using convex polytopes. Their location does not depend on the configuration. Therefore, no transformation of the vertices is necessary.

\subsection{Minimum Distance Equivalence}


The distance between two bodies $\nset B_1$ and $\nset B_2$ is defined by the optimization problem
\begin{mini}|s|
{\nvec w, \nvec x_1, \nvec x_2}{\left\lVert \nvec w \right\rVert}{}{\Delta(\nset B_1, \nset B_2)=}
\addConstraint{\nvec w}{= \nvec x_1 - \nvec x_2}
\addConstraint{\nvec x_1}{\in \nset B_1}
\addConstraint{\nvec x_2}{\in \nset B_2}.
\label{eq:mini_distance_bodies}
\end{mini}
This distance is not differentiable with respect to the pose of the bodies in many cases \cite{tracy2023a}. In order to use state-of-the-art NLP solvers, which make use of derivatives of constraint functions, a reformulation is therefore necessary.

In~\cite{dietz2023a}, this problem is overcome for the case of convex polytopes and ellipsoids. They prove the equivalence
\begin{equation}
\begin{aligned}
\Delta(\nset P_1, \nset P_2) \geq \Delta_\mathrm{min} \Leftrightarrow &\\
\exists \nvec \xi \in \mathbb R^{3},\ \lambda,\mu \in \mathbb R : -\frac{1}{4} \nvec\xi\transposed \nvec\xi - \lambda - \mu &\geq \Delta_\mathrm{min}^2,\\
\nmat V_1\transposed \nvec\xi + \lambda\nvec 1 &\geq \nvec 0,\\
-\nmat V_2\transposed \nvec\xi + \mu\nvec 1 &\geq \nvec 0
\end{aligned}
\label{eq:duality_equivalence}
\end{equation}
for the minimal distance $\Delta_\mathrm{min} > 0$ of two polytopes $\nset P_1,\nset P_2$ with vertex matrices $\nmat V_1$ and $\nmat V_2$.

Their proof is based on Lagrange duality. First, they reformulate the minimization problem~\eqref{eq:mini_distance_bodies} by using the squared Euclidean distance and substituting the polytope representation~\eqref{eq:polyeder_menge} into the constraints. By taking the dual problem of this minimization problem and simplifying, they obtain a maximization problem which, because of strong duality, yields the same distance. The decision variables of this dual problem are the newly introduced dual variables $\nvec\xi,\ \lambda$ and $\mu$, which are called collision multipliers. Using the dual maximization problem to obtain a lower bound on the actual distance, the inequalities in~\eqref{eq:duality_equivalence} are obtained.

\begin{figure}
    \centering
    \begin{subfigure}{0.5\columnwidth}
    \centering
    \begin{tikzpicture}[scale=0.6]
        \def\R{1}
        \def\l{3}
        \coordinate (a) at (0,0);
        \coordinate (b) at (\l,0);
        \draw[dashed] (a) -- (b) node[midway,below] {$\nset L$};
        \fill (a) circle (1pt);
        \fill (b) circle (1pt);
        \draw ($ (a) + (0,\R) $) -- ($ (b) + (0,\R) $);
        \draw ($ (a) + (0,-\R) $) -- ($ (b) + (0,-\R) $);
        \draw ($ (a) + (0,\R) $) arc (90:270:\R);
        \draw ($ (b) + (0,-\R) $) arc (-90:90:\R);
        \coordinate (c) at (2,3);
        \draw (c) -- (3,4) -- (0,4) -- (c);
        \fill (c) circle (1pt);
        \fill (3,4) circle (1pt);
        \fill (0,4) circle (1pt);
        \draw[stealth-stealth] ($(a) + (2,\R)$) -- (c) node[midway,right] {$\Delta_\mathrm{min}$};
        \draw[decorate,decoration={brace,amplitude=6pt}] ($(a) + (2,0)$) -- (c) node[midway,left,xshift=-6pt] {$\Delta_\mathrm{min} + r$};
        \draw[stealth-stealth] ($(a) + (2,0)$) -- ($(a) + (2,\R)$) node[midway,right] {$r$};
        \node at (c) [above,yshift=3pt] {$\nset P$};
    \end{tikzpicture}
    \caption{Capsule and polytope.}
    \label{fig:capsule_polytope_distance_illustration}
    \end{subfigure}%
    \begin{subfigure}{0.5\columnwidth}
    \centering
    \begin{tikzpicture}[scale=0.6]
        \def\R{1}
        \def\r{1.25}
        \def\l{3}
        \coordinate (a) at (0,0);
        \coordinate (b) at (\l,0);
        \fill (a) circle (1pt);
        \fill (b) circle (1pt);
        \draw[dashed] (a) -- (b) node[midway,below] {$\nset L_1$};
        \draw ($ (a) + (0,\R) $) -- ($ (b) + (0,\R) $);
        \draw ($ (a) + (0,-\R) $) -- ($ (b) + (0,-\R) $);
        \draw ($ (a) + (0,\R) $) arc (90:270:\R);
        \draw ($ (b) + (0,-\R) $) arc (-90:90:\R);
        \begin{scope}[shift={(2,3)}, rotate=150]
            \coordinate (c) at (0,0);
            \coordinate (d) at (\l,0);
            \draw[dashed] (c) -- (d) node[midway,below] {$\nset L_2$};
            \draw ($ (c) + (0,\r) $) -- ($ (d) + (0,\r) $);
            \draw ($ (c) + (0,-\r) $) -- ($ (d) + (0,-\r) $);
            \draw ($ (c) + (0,\r) $) arc (90:270:\r);
            \draw ($ (d) + (0,-\r) $) arc (-90:90:\r);
            \fill (c) circle (1pt);
            \fill (d) circle (1pt);
        \end{scope}
        \draw[stealth-stealth] (2,\R) -- ($ (c) + (0,-\r) $) node[midway,right] {$\Delta_\mathrm{min}$};
        \draw[decorate,decoration={brace,amplitude=6pt}]
        (2,0) -- (c)
        node[midway,left,xshift=-6pt] {$\Delta_\mathrm{min}+r_1+r_2$};
        \draw[stealth-stealth] (2,0) -- (2,\R) node[midway,right] {$r_1$};
        \draw[stealth-stealth] (c) -- ($ (c) + (0,-\r) $) node[midway,right] {$r_2$};
    \end{tikzpicture}
    \caption{Two capsules.}
    \label{fig:capsule_distance_illustration}
    \end{subfigure}
    \caption{Illustration of the minimal required distances.}
    \label{fig:Capsule_minimal_distances}
\end{figure}
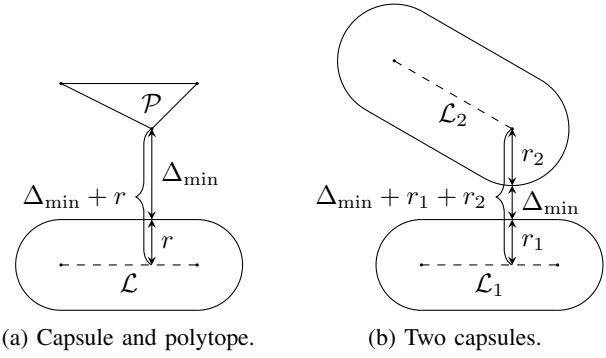

We adapt equivalence~\eqref{eq:duality_equivalence} for capsules by exploiting the fact that a capsule can be treated like a convex polytope by increasing the distance threshold by the radius of the capsule and exploiting the fact that the center line is a convex polytope, i.e.,
\begin{equation}
    \Delta(\nset P, \nset C) \geq \Delta_\mathrm{min} \Leftrightarrow \Delta(\nset P, \nset L) \geq \Delta_\mathrm{min} + r
    \label{eq:polytope_capsule_distance}
\end{equation}
for constraints concerning a convex polytope $\nset P$ and a capsule $\nset C$ (see \autoref{fig:capsule_polytope_distance_illustration}), and
\begin{equation}
    \Delta(\nset C_1, \nset C_2) \geq \Delta_\mathrm{min} \Leftrightarrow \Delta(\nset L_1, \nset L_2) \geq \Delta_\mathrm{min} + r_1 + r_2
    \label{eq:capsules_distance}
\end{equation}
in the case of two capsules $\nset C_1, \nset C_2$ with respective center lines $\nset L_1, \nset L_2$ and radii $r_1,r_2$ (see \autoref{fig:capsule_distance_illustration}).

Applying the equivalence~\eqref{eq:duality_equivalence} as collision-avoidance constraints into an OCP introduces five additional variables (collision multipliers) per pair of collision objects, together with five scalar inequalities for each capsule–capsule pair and $3+N_\mathrm{v}$ scalar inequalities for each capsule–polytope pair.

\section{Collision Free Trajectory Planning}
A key component of robotic control architectures is the generation of joint trajectories, which are then tracked by low-level controllers, typically PID schemes. Trajectory planning is commonly formulated as an optimal control problem (OCP), where an objective function $J$ is minimized over a finite prediction horizon with respect to the control input $\nvec{u}$. Knowledge of the physical model is incorporated in the shape of a differential equation enforced through equality constraints and additional inequality constraints on the states $\nvec{x}$ and inputs of the system can be added. For numerical implementation, the continuous-time problem is transcribed into a nonlinear program (NLP) under suitable assumptions. This work applies a multiple-shooting approach \cite{rawlings2022b} with the control input $\nvec{u}(t)$ defined as a piecewise constant function. The path constraints are thus only evaluated at the shooting nodes. The proposed NLP is constructed over the following subsections.

\subsection{Motion retargeting and input prediction}
In a teleoperation setting, the human input needs to be mapped from the VR controller space onto a desired pose of the end-effector in the robot space. This is done with a relative motion retargeting as described in \cite{grobbel2025c}. The user can move the VR controller freely, and only while activating a clutching mechanism, the operator's input is mapped onto a desired pose for the robot end-effector. The desired pose at time step $k$ in the robot space is then given with a homogeneous transformation matrix $\nmat{T}_{\mathrm{ref}}[k] \in \mathbb{R}^{4 \times 4}$ which contains the rotational matrix $\nmat{R}_{\mathrm{ref}}$ and the translation vector $\nvec{p}_{\mathrm{ref}}$.

For a predictive planning approach, a reference trajectory $\nmat T_\mathrm{ref}:\ \{1,\ldots,N\}\rightarrow \mathbb R^{4\times 4}$ over the prediction horizon needs to be generated. The reference starts from step $k=1$, since the robot states are fixed for the first step $k=0$. A widely spread approach is based on the assumption of a movement with constant velocity \cite{chipalkatty2013a} for translation and a constant angular velocity assumption around a constant rotation axis for rotational movements. 

\subsection{Robot Model}
A robot manipulator with $n_q$ rotational joints $\nvec{q} = [q_0\ \dots \ q_{n_q}]\transposed$ is considered. With the assumption of sufficiently performant low-level controllers, the linear state space model
\begin{equation}
    \nvec{x}[k+1] = \nmat{A}\nvec{x}[k] + \nmat{B}\nvec{u}[k],
\end{equation}
which is a double integrator, can be applied to predict the kinematic robot movements with the state transition matrix $\nmat{A}$ and the input mapping matrix $\nmat{B}$. The discrete-time transition matrix $\nmat{A}$ can be computed with the matrix exponential from the continuous system. The state vector $\nvec{x} = [\nvec{q}\transposed\ \dot{\nvec{q}}\transposed]\transposed$ consists of the concatenation of joint angles and velocities, and the input vector $\nvec{u} = \ddot{\nvec q}$ consists of the joint accelerations. $\nvec{x}$ and $\nvec{u}$ are interpreted as sequences over a prediction horizon of length $N$ with the respective elements $\nvec{x}[k] \in \mathbb{R}^{2n_q}$ at time step $k$ and $k \in \{0, \dots, N \}$, while $\nvec{u}[k] \in \mathbb{R}^{n_q}$ is only defined up to $k=N-1$.

\subsection{Objective function}
The main objective in a teleoperation scenario is the tracking of the given user input $\nmat{T}_{\mathrm{ref}}$. The current position $\nvec{p}(\nvec{q}[k])$ and orientation $\nmat{R}(\nvec{q}[k])$ of the end-effector are extracted from the homogeneous transformation matrix $\nmat{T}(\nvec{q}[k])$, which is given through the forward kinematics. 

The positional tracking error can then be stated as the squared weighted error
\begin{equation}
l_{\mathrm p}\left[k\right] = (\nvec{p}[k] -\nvec{p}_{ref}[k])\transposed \cdot \nmat{Q}_{\mathrm p} \cdot (\nvec{p}[k] -\nvec{p}_{ref}[k])
\end{equation}
with the diagonal matrix $\nmat{Q}_p$ containing the elements of the weight vector $\nvec{w}_{p} \in \mathbb{R}^3$ as entries. This allows the assignment of individual weights for the three translational degrees of freedom of the end-effector.

For rotational tracking, an approach based on the Frobenius norm \cite{huynh2009a} is adopted. The stage cost is formulated as
\begin{equation}
l_{\mathrm o}[k] = w_{\mathrm o} \cdot  \lVert \mathbb{I} - \nmat R[k] \nmat R_\mathrm{ref}[k]\transposed \rVert^2_F
\end{equation}
with the scalar weight $w_{\mathrm o}$. The multiplication of $\nmat R [k] \nmat R_\mathrm{ref}[k]\transposed$ becomes the identity matrix $\mathbb{I}$ under perfect alignment of the end-effector with the desired orientation, thus driving the cost $l_{\mathrm o}$ to 0. The square of the Frobenius norm is introduced here to prevent computational complexity by removing the square root operations.

A secondary set of objectives is introduced for the regularization of the optimization variables and the generation of smooth trajectories. The joint velocities $\dot{\nvec{q}}$ and the joint accelerations $\ddot{\nvec{q}} = \nvec{u}$ are considered under the quadratic cost terms
\begin{align}
    &l_{\mathrm a} [k] = \dot{\nvec{q}}[k]\transposed \cdot \nmat{Q}_{\mathrm a} \cdot \dot{\nvec{q}}[k]\\
    &l_{\mathrm b} [k] = \ddot{\nvec{q}}[k]\transposed \cdot \nmat{Q}_{\mathrm b} \cdot \ddot{\nvec{q}}[k]
\end{align}
with the diagonal weight matrices $\nmat{Q}_{\mathrm a}$ and $\nmat{Q}_{\mathrm b}$

Slack variables are a common method to relax inequality constraints and to keep the optimization problem feasible. This becomes especially important in real-world applications, since through disturbances on the real system or through numerical inaccuracies, the state $\nvec{x}_0$ might be slightly out of the allowed regions. This is especially likely when operating close to collision constraints, since the optimum of the optimization problem often lies on some boundaries. Slack variables push the trajectory planner back to feasible regions by a high cost.
A single slack variable $s[k]$ for every time step $k$ is included in the objective function as a third objective with the weight $w_{\mathrm s}$ as quadratic stage cost 
\begin{equation}
    l_{\mathrm s} [k] = w_{\mathrm s} \cdot s[k]^2.
\end{equation}

All stage costs $l_i$ are combined in the objective function as a sum of all stage costs over the prediction horizon
\begin{align} \label{eq:objective}
    J &= \sum_{k = 1}^{N} \underbrace{l_{ \mathrm o}[k] + l_{\mathrm p}[k]}_{\mathrm{tracking}} +  \underbrace{l_{ \mathrm a}[k] + l_{\mathrm b}[k-1]}_{\mathrm{regularization}} + \underbrace{l_{\mathrm s}[k]}_{\mathrm{slack}}.
\end{align}
Since the state variable is fixed through the starting constraint for the first time index $k=0$, the sum over all stage costs starts with time index $k=1$ and only the stage cost $l_{2b}$, containing the control input $\nvec{u}$, is starting from $k=0$, which is also only defined up to $k=N-1$. The final stage cost is thus integrated into the sum without an explicit set of weights.

\subsection{Collision Constraints}
Collision avoidance is included in the optimization problem as a set of inequality constraints. In the following, equations for self-collision and environmental collision avoidance are derived.

For self-collisions, a set $\nset I_\mathrm s \subset \{0,...,N_\mathrm s-1\}^2$ of relevant collision pairs is considered. It contains all tuples $(i,j)$ where collisions between capsules $\nset C_i$ and $\nset C_j$ of the robot collision model should be avoided. Some capsule pairs do not need to be considered for collision avoidance, since they kinematically cannot come into contact (e.g., shoulder and elbow).

For each relevant pair $(i,j)\in\nset I_\mathrm s$ and each time step $k\in \{1,\ldots,N\}$, the collision multipliers $\nvec\gamma_{\mathrm s,ij}[k] = [\nvec\xi_{\mathrm s,ij}[k]\transposed\ \lambda_{\mathrm s,ij}[k]\ \mu_{\mathrm s,ij}[k]]\transposed\in\mathbb R^5$ are introduced as decision variables, along with a constraint function
\begin{equation} \label{eq:self_collision}
\begin{aligned}
&\nvec g_{\mathrm s, ij} (\nvec q, \nvec\gamma_{\mathrm s,ij})\\
&= \begin{bmatrix}
\frac{1}{4} \nvec\xi_{\mathrm s,ij}\transposed \nvec\xi_{\mathrm s,ij} + \lambda_{\mathrm s,ij} + \mu_{\mathrm s,ij} + (\Delta_\mathrm{min} + r_i + r_j)^2\\
-\nmat V_{\nset L_i}(\nvec q)\transposed \nvec\xi_{\mathrm s,ij} - \lambda_{\mathrm s,ij} \nvec 1\\
\nmat V_{\nset L_j}(\nvec q)\transposed \nvec\xi_{\mathrm s,ij} - \mu_{\mathrm s,ij} \nvec 1\\
\end{bmatrix}.
\end{aligned}
\end{equation}
Here, $r_i, r_j$ are the radii of the capsules $\nset C_i$ and $\nset C_j$, respectively, and $V_{\nset L_i}(\nvec q),V_{\nset L_j}(\nvec q)\in\mathbb R^{3\times 2}$ are their vertex matrices containing the end points of their center lines as columns. These matrices depend on the configuration $\nvec q$ as given in~\eqref{eq:vertices_transformation}. The dependence on the time step $k$ is omitted for compactness.

With \autoref{eq:self_collision}, the inequality $\nvec g_{\mathrm s, ij} (\nvec q[k], \nvec\gamma_{\mathrm s,ij}[k]) \leq \nvec 0$ for all $k\in\{1,\ldots,N\}$ guarantees avoidance of self-collision. The interpretation is as follows: Every feasible point of the planning problem corresponds to a configuration trajectory $\nvec q$ for which the distance requirement $\Delta(\nset C_i, \nset C_j) \geq \Delta_\mathrm{min}$ is satisfied at all times, due to the equivalences~\eqref{eq:duality_equivalence} and~\eqref{eq:capsules_distance}. Conversely, these equivalences guarantee that for every configuration trajectory satisfying the distance requirement, there exist collision multipliers for which $\nvec g_{\mathrm s, ij} (\nvec q[k], \nvec\gamma_{\mathrm s,ij}[k]) \leq \nvec 0$ is satisfied, i.e., this inequality does not restrict the trajectory more than necessary to guarantee the distance requirement.

The avoidance of collisions with environmental obstacles is treated analogously by introducing the collision multipliers $\nvec\gamma_{\mathrm s,ij}[k] = [\nvec\xi_{\mathrm o,ij}[k]\transposed\ \lambda_{\mathrm o,ij}[k]\ \mu_{\mathrm o,ij}[k]]\transposed\in\mathbb R^5$ and the constraint function
\begin{equation}
\begin{aligned}
&\nvec g_{\mathrm o, ij} (\nvec q, \nvec\gamma_{\mathrm o,ij})\\
&= \begin{bmatrix}
\frac{1}{4} \nvec\xi_{\mathrm o,ij}\transposed \nvec\xi_{\mathrm o,ij} + \lambda_{\mathrm o,ij} + \mu_{\mathrm o,ij} + (\Delta_\mathrm{min} + r_i)^2\\
-\nmat V_{\nset L_i}(\nvec q)\transposed \nvec\xi_{\mathrm o,ij} - \lambda_{\mathrm o,ij} \nvec 1\\
\nmat V_{j}\transposed \nvec\xi_{\mathrm o,ij} - \mu_{\mathrm o,ij} \nvec 1\\
\end{bmatrix}
\end{aligned}
\end{equation}
for every $(i,j)\in\nset I_\mathrm o \subset \{0,\ldots,N_\mathrm s-1\}\times \{0,\ldots,N_\mathrm o-1\}$ that corresponds to a relevant pair of a capsule $\nset C_i$ of the robot model and a convex environmental obstacle polytope $\nset P_j$ with vertex matrix $\nmat V_j$.

To improve efficiency, the surface of the table that the manipulator is mounted on is not modeled as a polytope, but instead as a horizontal plane that all capsules need to stay above. The corresponding constraint functions are given by
\begin{equation}
    \nvec g_{\mathrm t,i}(\nvec q)= -([0\ 0\ 1] \nmat V(\nvec q) )\transposed + (\Delta_\mathrm{min} + r_i)\nvec 1
\end{equation}
for each $i \in \{0,\ldots,N_\mathrm s -1\}$, requiring no collision multipliers.

The entire collision constraint function $\nvec g(\nvec x, \nvec\gamma)$ is obtained by vertically concatenating all $\nvec g_{\mathrm s,ij}$ with $(i,j) \in\nset I_\mathrm s$, all $\nvec g_{\mathrm o,ij}$ with $(i,j) \in\nset I_\mathrm o$, and all $\nvec g_{\mathrm t,i}$ with $i \in \{0,\ldots,N_\mathrm s -1\}$, with $\nvec\gamma$ being the vector of all collision multipliers, which is also constructed by vertical concatenation.

\subsection{Nonlinear Optimization Problem}
The objective function \eqref{eq:objective} is minimized with respect to the system states $\nvec x$, system input $\nvec u$, the slack variables $s$ and the collision multipliers $\nvec \gamma$ over the prediction horizon, resulting in the NLP
\begin{align} \label{eq:NLP}
    &\underset{\nvec{x}, \nvec{u}, \nvec{\gamma}, s}{\min}  && J \left( \nvec{x}, \nvec{u}, s, \nmat{T}_{\mathrm{ref}} \right) \\
    &\text{s.t.} \nonumber \\
    &&& \nvec{x}[k+1] = \nmat{A} \nvec{x}[k] + \nmat{B} \nvec{u}[k]  & \forall k \in \{0,\dots, N - 1\} \nonumber\\
    &&& \nvec{x}_{\min} \leq \nvec{x}[k] \leq \nvec{x}_{\max} & \forall k  \in \{1,\dots,  N \} \nonumber\\
    &&& \nvec{u}_{\min} \leq \nvec{u}[k] \leq \nvec{u}_{\max} & \forall k \in  \{0,\dots,  N - 1\} \nonumber\\
    &&& \nvec{g}(\nvec{x}[k],  \nvec{\gamma}[k]) - s[k] \nvec 1 \leq \nvec{0} & \forall k \in \{1,\dots, N\}\nonumber \\
    &&& s[k] \geq 0 & \forall k \in \{1,\dots, N\}\nonumber \\
    &&& \nvec{x}[k=0]  = \nvec{x}_0 \nonumber      .
\end{align}

The slack variable $s[k]$ is subtracted from the inequality constraint $\nvec g$. While $s[k]$ is also constrained to always be positive, this allows single entries of $\nvec g[k]$ to become positive, violating the collision constraints, at a high cost in the objective function. To reduce the number of optimization variables, only one slack variable is defined and applied to all inequality constraints for each time step $k$. 

In addition to the collision constraints, minimal and maximal values are imposed on joint angles $\nvec q$, angular velocities $\dot{\nvec q}$, and angular accelerations $\nvec u$. The collision multipliers $\nvec \gamma$ don't appear in the objective function, but still need to be considered by the applied solver to find collision-free trajectories. 

\section{Experimental Setup}
We evaluate our proposed method for collision avoidant trajectory planning in teleoperation on a UR5e manipulator. First it is compared against the two common approaches with sphere approximation \cite{rubagotti2019a} and witness points \cite{hu2021a} in a simulation with an increasing number of obstacles, to test for scalability. Secondly, we test our method in a real teleoperation setting by trying to deliberately cause collisions. We check for computation time, distance to obstacles and tracking accuracy. 

The NLP \autoref{eq:NLP} is implemented with Python and CasADi \cite{andersson2019} and the open source NLP solver IPOPT \cite{wachter2009a}, and solved at a fixed sample rate of 20 Hz for a given reference trajectory $\nmat T_{\mathrm{ref}}$ and the current state $\nvec x_0$. The CasADi built-in just-in-time compilation has been utilized with the highest optimization level, so that all computationally demanding tasks are run as compiled binaries. The utilized hardware is an AMD Ryzen 7950X CPU. The implementation is publicly available at: https://github.com/GiantMGG/robot-optimal-trajectory-planning.

\subsection{Implementation of Baseline Approaches}
The first baseline method \cite{hu2021a} is implemented with several modifications. As the original work does not fully describe the quadratic approximation of the objective, and since a nonlinear solver was used, we employ IPOPT without quadratic approximation. This yields reliable convergence at the cost of higher computation times. To ensure recursive feasibility between SQP iterations, slack variables are added to the collision constraints and penalized quadratically in the objective. The UR5e is approximated with capsules, consistent with our approach. For a fair comparison, the reference is adapted from a constant setpoint to a constant-velocity prediction.

The second baseline, based on sphere approximations (e.g., \cite{rubagotti2019a}), is directly integrated into the optimization problem \autoref{eq:NLP} and aligned with the capsule formulation for comparability. The robot is represented by 29 spheres distributed along the centerlines of the corresponding capsules. Collision constraints for both self- and environment interactions are expressed as distance functions between relevant sphere pairs, using the sum of their radii. Environmental obstacles are modeled with three spheres in simulation and five spheres in the real-robot experiments.

\subsection{Introduction of Testcases}\label{sec:testcases_intro}
All three systems are evaluated in a simplified simulation, where the robot state is assumed to follow the planned trajectories exactly, without dynamics or disturbances. Three 20–30 s scenarios with pre-recorded references are defined: (i) free arm jogging, (ii) end-effector motion into the robot base to trigger self-collisions, and (iii) motion to trigger collisions with the environment. Each scenario is varied by the number of cuboid obstacles (0, 1, 3, 6, 9, 12). The obstacles are either modeled as polytopes with 8 vertices, or with 3 spheres. To reduce peripheral effects on computation time, each configuration is run four times. This test case analyzes the scalability of the approaches with increasing number of collision constraints. 

A second test case for evaluation is run on an UR5e robot. A larger cuboid, modeled with 5 spheres or polytope with 8 vertices, serves as a single obstacle (\autoref{fig:capsule_traj_3d}). The operator's task is to try to touch the obstacle with parts of the robot from multiple sides and angles and to induce self-colliding motion. This testcase shows the applicability to a real robot in a teleoperation scenario, which closes the overall control loop including the human operator, allowing for analysis of the tracking and collision avoidance performance.

\section{Experimental Results and Discussion}
\begin{table*}[t]
\centering
\caption{Computation times (ms) for three scenarios with varying numbers of obstacles, averaged over four runs.}
\label{tab:comp_times}
\setlength{\tabcolsep}{4pt}
\renewcommand{\arraystretch}{1.0}
\begin{tabular}{c|c|cc|cc|cc|cc|cc|cc}
Scenario 
 & Method 
 & \multicolumn{2}{c|}{0 Obstacles} 
 & \multicolumn{2}{c|}{1 Obstacle} 
 & \multicolumn{2}{c|}{3 Obstacles} 
 & \multicolumn{2}{c|}{6 Obstacles} 
 & \multicolumn{2}{c|}{9 Obstacles} 
 & \multicolumn{2}{c}{12 Obstacles}  \\
 \cline{3-4}\cline{5-6}\cline{7-8}\cline{9-10}\cline{11-12}\cline{13-14}
 & & mean$\pm$std & max & mean$\pm$std & max & mean$\pm$std & max & mean$\pm$std & max & mean$\pm$std & max & mean$\pm$std & max \\
\hline
(i) Arm jogging & NMPC-MP \cite{hu2021a} & 151$\pm$0.3 & 168 & 195$\pm$0.3 & 210 & 301$\pm$0.3 & 326 & 453$\pm$2.0 & 477 & 623$\pm$2.7 & 654 & 797$\pm$3.1 & 849 \\
 & Ours & 15$\pm$0.3 & 25 & 22$\pm$0.3 & 49 & 46$\pm$0.5 & 76 & 114$\pm$0.4 & 186 & 196$\pm$0.9 & 493 & 316$\pm$1.3 & 552 \\
 & Spheres \cite{rubagotti2019a} & 20$\pm$0.3 & 36 & 29$\pm$0.3 & 46 & 58$\pm$1.1 & 99 & 116$\pm$1.4 & 170 & 202$\pm$1.1 & 269 & 301$\pm$0.8 & 392 \\
 \hline
(ii) Self-collision & NMPC-MP \cite{hu2021a}& 155$\pm$0.9 & 198 & 195$\pm$0.0 & 245 & 299$\pm$1.3 & 360 & 467$\pm$2.6 & 544 & 640$\pm$2.0 & 739 & 809$\pm$4.5 & 1006 \\
 & Ours & 14$\pm$0.0 & 28 & 24$\pm$0.3 & 39 & 62$\pm$0.6 & 127 & 107$\pm$0.8 & 275 & 210$\pm$1.8 & 345 & 398$\pm$2.6 & 2884 \\
 & Spheres \cite{rubagotti2019a}& 25$\pm$0.0 & 49 & 34$\pm$0.3 & 57 & 84$\pm$0.9 & 206 & 130$\pm$0.9 & 214 & 305$\pm$2.1 & 1639 & 448$\pm$1.9 & 2481 \\
 \hline
(iii) Obstacles & NMPC-MP \cite{hu2021a}& 166$\pm$0.0 & 197 & 209$\pm$0.5 & 258 & 314$\pm$0.9 & 352 & 488$\pm$1.6 & 593 & 681$\pm$3.4 & 759 & 881$\pm$3.0 & 1011 \\
 & Ours & 14$\pm$0.3 & 22 & 26$\pm$0.0 & 40 & 59$\pm$0.9 & 89 & 124$\pm$1.2 & 250 & 231$\pm$0.8 & 664 & 377$\pm$2.8 & 850 \\
 & Spheres \cite{rubagotti2019a}& 21$\pm$0.3 & 37 & 36$\pm$0.4 & 55 & 75$\pm$1.1 & 117 & 148$\pm$1.3 & 232 & 248$\pm$0.4 & 347 & 366$\pm$1.2 & 534 \\
\end{tabular}
\vspace{-5pt}
\end{table*}

\subsection{Evaluation in Simulation}
The results of the simulative study, with emphasize on computation time, are summarized in \autoref{tab:comp_times}. The free arm jogging scenario is consistently solved with the lowest computation times across all methods, indicating that the other two scenarios indeed activate the collision constraints and therefore impose higher computational demand.

Among the evaluated methods, NMPC-MP exhibits the highest computation time but shows better scalability with the number of obstacles. The original publication employed a quadratic programming (QP) solver, which may reduce computation times considerably; however, we were unable to achieve reliable convergence in our implementation. This may be attributed to the original Schulman approximation \cite{schulman2013a}, which assumes one witness point to remain fixed in space, an assumption that does not hold in the context of self-collision avoidance. 
 
The sphere-based approach and our collision-multiplier formulation yield comparable results, with our method showing slightly lower mean and maximum computation times. The resulting joint trajectories are also very similar across both methods. In contrast, NMPC-MP produces less smooth trajectories, since its system model does not account for joint accelerations.

With a prediction horizon of $N=10$ steps, the sphere-based approach already struggles in the presence of a single obstacle, whereas our approach still maintains the cycle time of 50 ms. It should also be noted that the approximation of the cuboids with three spheres is rather inaccurate.
 
Neither the sphere-based method nor our approach scales well with an increasing number of obstacles and thus cannot be applied in real-time teleoperation with a horizon of $N=10$ in cluttered environments. For every new obstacle and each capsule of the robot model, our approach introduces 11 additional constraints and 5 optimization variables per time step. Depending on the robot link (some approximated by up to 7 spheres), as many as 21 constraints may be added, since each cuboid is approximated by three spheres.

\subsection{Evaluation on Real Manipulator}

\begin{figure*}[htbp] 
    \vspace{-31pt}
    \centering
    \begin{subfigure}[t]{0.3\textwidth}
        \centering
        \includegraphics[width=\textwidth]{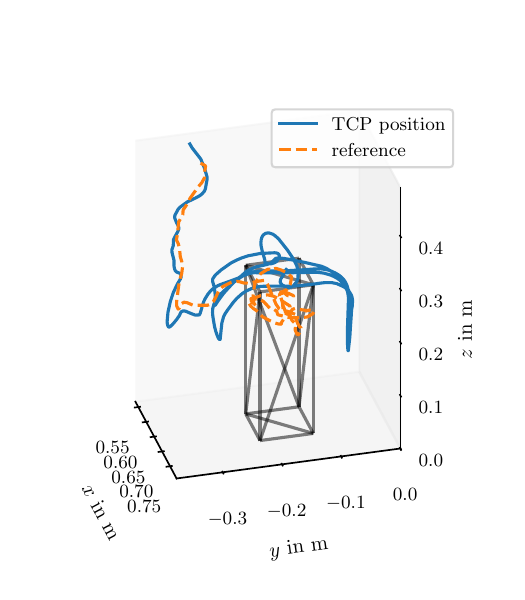}
        \caption{Trajectory of end-effector position.}
        \label{fig:capsule_traj_3d}
    \end{subfigure}%
    \hfill
    \begin{subfigure}[t]{0.345\textwidth}
        \centering
        \includegraphics[width=\textwidth]{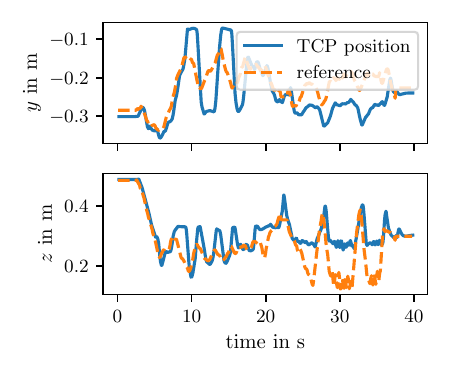}
        \caption{End-effector position in $y$ and $z$ direction.}
        \label{fig:capsule_traj_tracking}
    \end{subfigure}%
    \hfill
    \begin{subfigure}[t]{0.345\textwidth}
        \centering
        \includegraphics[width=\textwidth]{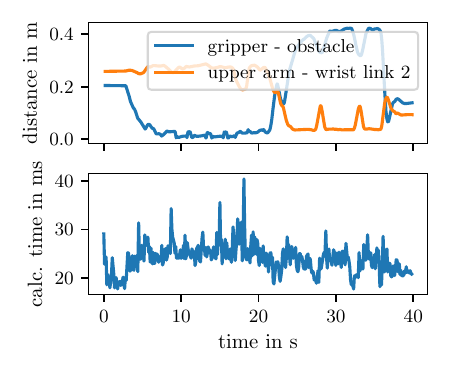}
        \caption{Distances and calculation time.}
        \label{fig:capsule_distances_and_calctime}
    \end{subfigure}
    \caption{Trajectory recording with our proposed method.}
    \label{fig:plots_capsule_traj}
    \vspace{-5pt}
\end{figure*}

\begin{figure*}[htbp] 
    \centering
    \begin{subfigure}[t]{0.3\textwidth}
        \centering
        \includegraphics[width=\textwidth]{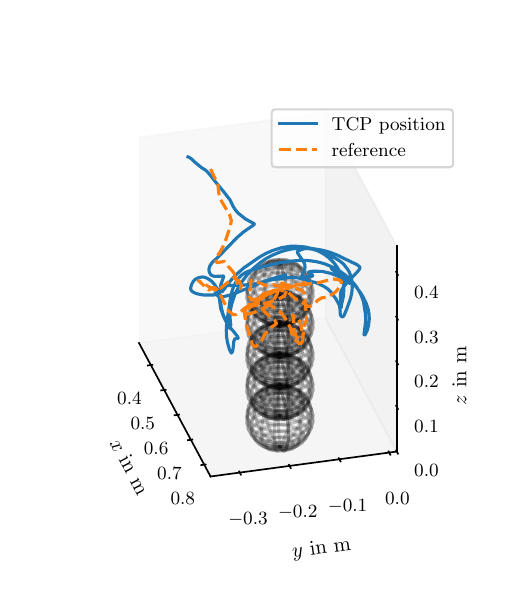}
        \caption{Trajectory of end-effector position.}
        \label{fig:spheres_traj_3d}
    \end{subfigure}%
    \hfill
    \begin{subfigure}[t]{0.345\textwidth}
        \centering
        \includegraphics[width=\textwidth]{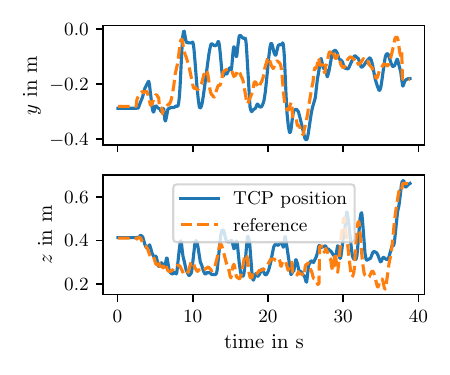}
        \caption{End-effector position in $y$ and $z$ direction.}
        \label{fig:spheres_traj_tracking}
    \end{subfigure}%
    \hfill
    \begin{subfigure}[t]{0.345\textwidth}
        \centering
        \includegraphics[width=\textwidth]{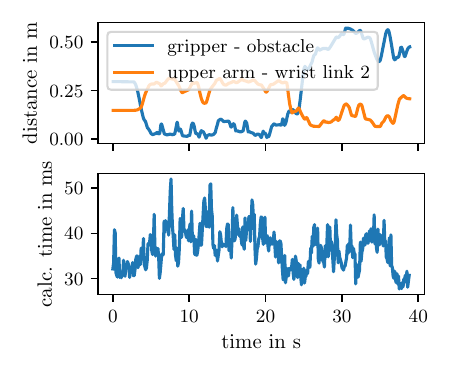}
        \caption{Distances and calculation time.}
        \label{fig:spheres_distances_and_calctime}
    \end{subfigure}
    \caption{Trajectory recording with robot and obstacle approximation through spheres.}
    \label{fig:plots_spheres_traj}
    \vspace{-0pt}
\end{figure*}

The collision-avoiding behavior of the planner is shown in \autoref{fig:plots_capsule_traj} and \autoref{fig:plots_spheres_traj}. The trade-off between tracking performance and collision avoidance appears in \autoref{fig:capsule_traj_tracking} and \autoref{fig:spheres_traj_tracking}, and is also evident in the 3D plots (\autoref{fig:capsule_traj_3d}, \autoref{fig:spheres_traj_3d}), where the operator-driven reference passes through the obstacle while the end-effector remains collision-free. The upper plots in \autoref{fig:capsule_distances_and_calctime} and \autoref{fig:spheres_distances_and_calctime} show the distances between gripper and obstacle (first 20 s), as well as between two links driven close together during teleoperation (last 20 s). These distances, computed from exact meshes, confirm that no collisions occur. A video of the experimental evaluation is available at: https://youtu.be/OR5T9Hy2fdE.

Compared to the sphere-based approach, the more accurate geometric modeling of robot and obstacles allows the manipulator to move closer to obstacles, reducing unnecessary evasive maneuvers (\autoref{fig:capsule_traj_3d} vs. \autoref{fig:spheres_traj_3d}). For computation time, our method remains between \SI{20}{\milli\second} and \SI{30}{\milli\second} for most of the trajectory, with peaks up to \SI{40}{\milli\second} during obstacle avoidance—still below the control cycle of \SI{50}{\milli\second} (see lower plot of \autoref{fig:capsule_distances_and_calctime}). The sphere-based approach exhibits longer runtimes, with two peaks exceeding \SI{50}{\milli\second}. In both methods, computation time increases near obstacles or self-collisions, when avoidance is required.





\section{Conclusion}
We proposed a method that combines differentiable collision constraints with accurate geometric modeling using capsules and polytopes. The approach achieves lower computation times than state-of-the-art methods while providing more precise obstacle representations. We validated the method in simulation with varying obstacle densities and on a UR5e manipulator in a real-world teleoperation study, demonstrating smoother and collision-free control. Although developed for teleoperation, the approach is directly transferable to general robotic applications. Future work will explore SQP-based methods to further reduce computation time.





\bibliographystyle{IEEEtran} 
\bibliography{99_references.tex}


\end{document}